\documentclass[10pt,journal]{IEEEtran}
\usepackage{lineno,bm,amsfonts,amsmath,algorithm,algorithmic,color,amssymb,multirow,amsthm,cite,booktabs,subfigure, } 
\theoremstyle{remark}
\theoremstyle{definition}

\ifCLASSINFOpdf
  \usepackage[pdftex]{graphicx}
  \graphicspath{{../pdf/}{../jpeg/}}
  \DeclareGraphicsExtensions{.pdf,.jpeg,.png}
\else
  \usepackage[dvips]{graphicx}
  \graphicspath{{../eps/}}
  \DeclareGraphicsExtensions{.eps}
\fi

\usepackage{epstopdf}
\begin{document}
%

\title{DeepFPC: Deep Unfolding of a Fixed-Point Continuation Algorithm for Sparse Signal Recovery from Quantized Measurements}
%
%
\author{Peng Xiao,
        Bin Liao,
        Nikos Deligiannis

\thanks{P. Xiao is with the College of
Electronics and Information Engineering, Shenzhen University, Shenzhen,
518060 P. R. China, and with the Department of Electronics
and Informatics, Vrije Universiteit Brussel, Brussels, Belgium (e-mail: peng.xiao@szu.edu.cn).}
\thanks{B. Liao is with the College of
Electronics and Information Engineering, Shenzhen University (e-mail: binliao@szu.edu.cn).}
\thanks{ N. Deligiannis is with the Department of Electronics
and Informatics, Vrije Universiteit Brussel, Brussels, Belgium, and with imec, Kapeldreef 75, B- 3001, Leuven, Belgium. (e-mail: ndeligia@etrovub.be).}}
\maketitle


\begin{abstract}
We present DeepFPC, a novel deep neural network designed by unfolding the iterations of the fixed-point continuation algorithm with one-sided $\ell_1$-norm (FPC-$\ell_1$), which has been proposed for solving the 1-bit compressed sensing problem. The network architecture resembles that of deep residual learning and incorporates prior knowledge about the signal structure (i.e., sparsity), thereby offering interpretability by design. Once DeepFPC is properly trained, a sparse signal can be recovered fast and accurately from quantized measurements. The proposed model is evaluated in the task of direction-of-arrival (DOA) estimation and is shown to outperform state-of-the-art algorithms, namely, the iterative FPC-$\ell_1$ algorithm and the 1-bit MUSIC method.
\end{abstract}

%
\IEEEpeerreviewmaketitle
\vspace{-1em}
\section{Introduction}
%
%
%
%

\IEEEPARstart{S}{parse} representation and compressed sensing (CS)~\cite{Donoho2006Compressed,Cand2008The} have led to good performance in various applications, including image processing\cite{Duarte2008Single,Wakin2007An,Lustig2008Compressed}, wireless communications\cite{Bajwa2006Compressive,Giannakis2010Distributed}, and direction-of-arrival (DOA) estimation\cite{malioutov2005sparse,6375850,7782330}. A variety of methods have been proposed to solve the related reconstruction problem, such as matching pursuit\cite{tropp2007signal,dai2009subspace}, basis pursuit\cite{chen1998atomic}, non-convex optimization\cite{Cand2008Enhancing,8747470}, and Bayesian algorithms\cite{4524050}. A more recent approach follows the paradigm of deep learning, which has shown success in many inference problems. Nevertheless, deep neural networks (DNNs) are considered black boxes---as their inner processes and generalization capabilities are not fully understood---and do not incorporate prior knowledge about the signal structure.

Deep unfolding\cite{gregor2010learning} promises to bridge the gap between analytical and deep-learning-based methods by designing DNNs as unrolled iterations of optimization algorithms. By incorporating knowledge about signal priors by design, deep unfolded networks offer interpretability compared to conventional DNNs, and have been shown to outperform traditional optimization-based methods and DNN models. Examples of such networks include LISTA~\cite{gregor2010learning}, LAMP~\cite{7934066}, ADMM-Net~\cite{sun2016deep}, $\ell_1$-$\ell_1$-RNN~\cite{Le2019Designing}, iRestNet~\cite{bertocchi2019deep} and  LeSITA~\cite{8844082}.

Following this research line, we propose a novel deep unfolded network, dubbed DeepFPC, to perform sparse signal recovery from quantized measurements. Each layer of DeepFPC corresponds to an iteration of the fixed point continuation algorithm with one-sided $\ell_1$-norm minimization (FPC-$\ell_1$), which is among the best-performing sparsity-free algorithms (i.e., it does not require knowing the sparsity level) for solving the 1-bit CS problem\cite{XIAO2019168}. Our contribution is as follows: (\textit{i}) we design a novel DNN that unfolds an iterative algorithm for sparse approximation from 1-bit measurements. Interestingly, the resulting network architecture resembles that of deep residual learning, implying that the model has the potential to be deeply trained. (\textit{ii}) We show that DeepFPC offers higher signal reconstruction accuracy and speed than the iterative FPC-$\ell_1$ algorithm. (\textit{iii}) We apply DeepFPC in the DOA estimation problem and show that it outperforms the state-of-the-art FPC-$\ell_1$\cite{XIAO2019168} and 1-bit MUSIC\cite{8700277} methods.

In the rest of the letter, Section~\ref{sec:background} reviews the background and, Sections~\ref{sec:ProposedModel}~and~\ref{sec:DOAestimation} present the proposed model and its  application in DOA estimation, respectively. Section~\ref{sec:Experiments} reports the experiments and Section~\ref{sec:conclusion} draws the conclusion.
\vspace{-0.8em}

\section{Background on the FPC-$\ell_1$ Method}
\label{sec:background}
In 1-bit CS, the task is to recover a signal from quantized measurements,
\begin{equation}\label{1}
\mathbf{y}=\operatorname{sign}(\mathbf{\Phi} \mathbf{x}+\bm{\varepsilon}),
\end{equation}
where $\mathbf{y}=\left[y_{1}\: y_{2}\ldots y_{M}\right]^{T} \in\{-1,1\}^{M}$ is the binary measurement vector, ${\bf\Phi} \in \mathbb{R}^{M \times N}$ is the sampling matrix, $\mathbf{x} \in \mathbb{R}^{N}$ is the sparse vector with $K\ll N$ nonzero entries, $\bm{\varepsilon}\in\mathbb{R}^{M}$ is a noise component, and $\operatorname{sign}(\cdot)$ is the sign function [$\operatorname{sign}(z)=1\:\text{if}\:z>0,\:\text{and}\:-1\:\text{otherwise}$]. A well-established algorithm for 1-bit CS recovery solves the following problem using the fixed-point continuation (FPC) method\cite{Boufounos20081}:
\begin{equation}\label{2}
\min _{\mathbf{x}}{\|{\bf x}\|_1+\lambda \sum_{i=1}^{M} h\left([\mathbf{Y} \mathbf{\Phi} \mathbf{x}]_{i}\right)}
\:{\text { s.t. }}\: {\|\mathbf{x}\|_{2}=1},
\end{equation}
where $\lambda>0$, $\mathbf{Y}=\operatorname{diag}(\mathbf{y})$, and $\sum_{i=1}^{M} h\left([\mathbf{Y} \mathbf{\Phi} \mathbf{x}]_{i}\right)$ is the consistent term, which can be specified as the one-sided $\ell_1$-norm, i.e., $h(z) :=\max \{0,-z\}$, or the one-sided $\ell_2$-norm, i.e., $h(z) :=\left(\max \{0,-z\}\right)^{2}$. Since the signal magnitude has been lost during the 1-bit quantization process, the signal is normalized with the unit $\ell_2$-norm
($\|{\bf x}\|_2=1$) to avoid a trivial solution. It is shown that the one-sided $\ell_1$-norm results in better performance than the one-sided $\ell_2$-norm\cite{XIAO2019168}; hence, the FPC-$\ell_1$ algorithm will be considered in what follows.

The FPC method \cite{Elaine2008} states that a way to solve Problem~\eqref{2} is to iteratively update $\bf x$ as
\begin{subequations}\label{4}
\begin{align}
\label{4-1}
& {\bf u}=S_{\nu}\left(\mathbf{x}^{(r)}-\tau\mathbf{g}(\mathbf{x}^{(r)})\right),
\\
\label{4-2}
& \mathbf{x}^{(r+1)}={{\bf{u}}}/{\| {\bf{u}}\|_2},
\end{align}
\end{subequations}
where $\tau>0$, $\nu=\tau/\lambda$, $\mathbf{g}({\bf x})$ is the gradient of the consistent term, and $S_{\nu}(\cdot) \triangleq \operatorname{sign}(\cdot) \odot \max \{|\cdot|-\nu, 0\}$ is the soft-thresholding operator. If the consistent term is specified as the one-sided $\ell_1$-norm, the gradient $\mathbf{g}(\mathbf{x})$ can be computed as
\begin{equation}\label{Gradient}
{{\bf g}}({\bf x})={\bf\Phi}^T\left(\operatorname{sign}\left({\bf\Phi x}\right)-{\bf y}\right).
\end{equation}

Consequently, the sparse signal can be reconstructed by the method with proper values for $\tau$, $\lambda$ and initialization $\mathbf{x}^{(0)}$.

\section{The Proposed DeepFPC Model}
\label{sec:ProposedModel}

We now present our DeepFPC architecture, which stems from unfolding the FPC-$\ell_1$ updates in~\eqref{4}, and describe how it should be trained to achieve high reconstruction
performance.


\subsection{The Deep Unfolded Residual Network Architecture}
\label{sec:NetArchitecture}
Replacing the gradient expression of \eqref{Gradient} in \eqref{4-1}, the latter can be written as
\begin{equation}\label{5}
\mathbf{u}=S_{\nu}\left(\mathbf{x}^{(r)}+{\bf C}\operatorname{sign}({\bf B} \mathbf{x}^{(r)})+{\bf A}\mathbf{y}\right),
\end{equation}
where ${\bf A}=\tau{\bf\Phi}^{T}$, ${\bf B}={\bf\Phi}$ and ${\bf C}=-\tau{\bf\Phi}^{T}$. Following the principles in\cite{gregor2010learning}, we can write $R$ iterations of the update rule defined by~\eqref{5} and \eqref{4-2} in the form of neural-network layers. The inputs to the network are an initial estimate of the signal~$\mathbf{x}^{(0)}$ and the (binary) measurement vector~$\mathbf{y}$, which corresponds to the bias term in DNN architectures. The matrices $\bf A$, $\bf B$, $\bf C$ are trainable linear weights that may vary per layer, and $S_{\nu}(\cdot)$ and $\operatorname{sign}(\cdot)$ play the role of element-wise non-linear activation functions.

We have empirically observed that the normalization step in \eqref{4-2} is not required to be applied after every layer; in effect, we have found that when normalization is performed after the final layer, the recovery performance is higher (we refer to Section \ref{sec:SignalReconstructionExperimentalResults} for details). We conjecture that since the network is trained with signal examples, it does not learn to produce the zero solutions with finite layers. Furthermore, since the $\textnormal{sign}(u)$ function is non-differential we approximate it using the hyperbolic tangent function $\textnormal{tanh}(\kappa u)=\frac{e^{\kappa u}-e^{-\kappa u}}{e^{\kappa u}+e^{-\kappa u}}$ with $\kappa\rightarrow\infty$. Hence, each layer in the proposed DeepFPC updates the signal estimate as $\mathbf{x}^{(r+1)}=S_{\nu}\left(\mathbf{x}^{(r)}+{\bf C}\operatorname{tanh}({\kappa\bf B} \mathbf{x}^{(r)})+{\bf A}\mathbf{y}\right)$, whereas the last layer normalizes the final estimate ${\mathbf{x}}^{\star}=\mathbf{x}^{(R)}/{\| \mathbf{x}^{(R)}\|_2}$.

\begin{figure}
  \centering
  \includegraphics[width=3.5 in]{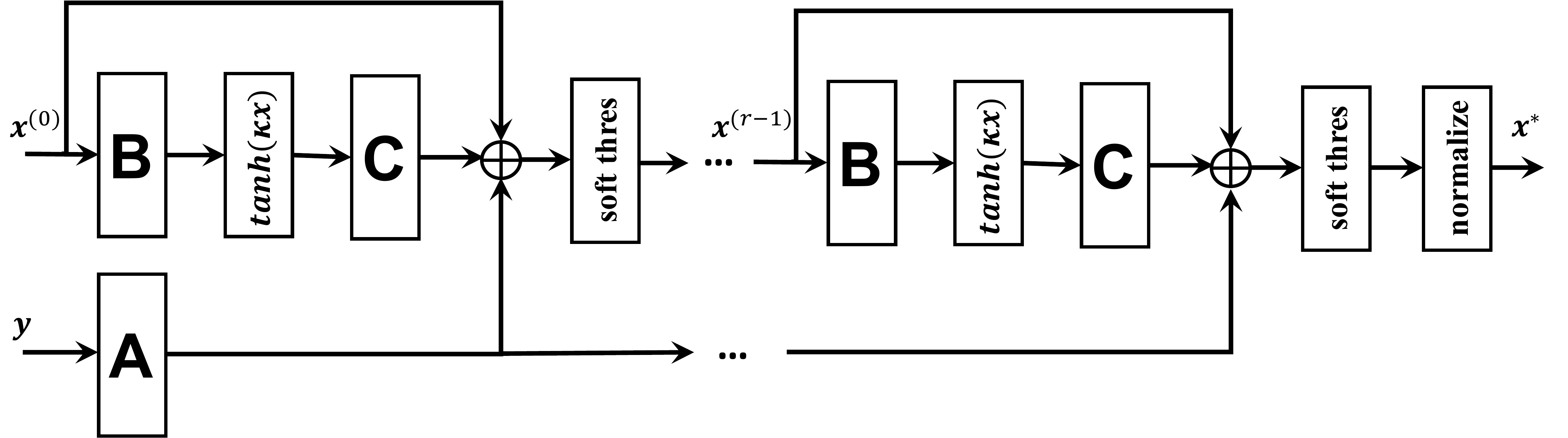}
  \caption{The block diagram of the proposed DeepFPC.}\label{Fig2}\vspace{-1.2em}
  \label{fig:NetworkArchitecture}
\end{figure}

The block diagram of DeepFPC is depicted in Fig. \ref{fig:NetworkArchitecture}. Interestingly, the network applies \textit{residual learning by design}; specifically, at each layer, the input to the soft-thresholding non-linearity is fed with a shortcut connection from the previous layer output. This makes the DeepFPC network similar to the ResNet architecture \cite{he2016deep} and implies that DeepFPC has the potential to be deeply trained.

\subsection{Training Strategies}
The trainable parameters per layer $r=1,2,\dots,R$ of DeepFPC are the weight matrices $\mathbf{A}_r$, $\mathbf{B}_r$, $\mathbf{C}_r$ and the soft-thresholding threshold $\nu_r$. These parameters can either vary per layer (untied weights strategy) or be shared over all layers (tied weights strategy). Experimentation showed that allowing $\nu_r$ to vary per layer leads to much better performance. On the contrary, allowing $\mathbf{A}_r$, $\mathbf{B}_r$ and $\mathbf{C}_r$ to be untied across layers does not notably improve performance and inflicts a significant memory overhead. Therefore, in order to maintain the memory complexity, we choose to train $\nu_r$ independently per layer, and share the weight-matrices $\bf A$, $\bf B$ and $\bf C$ over all layers; we refer to Section~\ref{sec:SignalReconstructionExperimentalResults} for further specifics.  We optimize the trainable parameter set ${\bf\Xi}=\{{\bf A,B,C,\bm\nu}\}$, with ${\bm \nu}=\{\nu_1,\nu_2,\dots,\nu_r,\dots,\nu_R\}$, using the training data set $\{{\bf y}^d,{\bf x}^d\}_{d=1}^D$ by minimizing a quadratic loss function as
\begin{equation}\label{6}
\mathcal{L}({\bf\Xi})=\frac{1}{D} \sum_{d=1}^{D}\left\|{\bf x}^\star\left({\bf y}^{d} ; {\bf\Xi}\right)-{\bf x}^{d}\right\|_{2}^{2},
\end{equation}
where $D$ is the batch size. The following remarks are helpful when training the DeepFPC model.

{\emph{Remark 1:}} 1-bit CS targets reducing the number of bits per sample rather than decreasing the number of samples; hence, $\bf \Phi$ in \eqref{1} can have more rows than columns, i.e., $M > N$. This means that the size of the weight matrices in DeepFPC is typically larger than those in alternative deep-unfolded networks for signal recovery such as LISTA\cite{gregor2010learning} and  LeSITA\cite{8844082}. Therefore, in order to avoid converging to undesired local minima, we adhere to the training strategy reported in \cite{vincent2010stacked,7934066}. Specifically, the parameters are not learned for all layers directly; instead, we successively learn ${\bf\Xi}^{(1)}, {\bf\Xi}^{(2)},...,{\bf\Xi}^{(R)}$, where ${\bf\Xi}^{(r)}$ represents the parameters of all layers up to $r$, i.e., ${\bf\Xi}^{(r)}=\{{\bf A,B,C},\{\nu_t\}_{t=1}^{r}\}$. Moreover, when training ${\bf\Xi}^{(r)}$, we first initialize $\nu_r$ with $\nu_{r-1}$ and learn $\nu_{r}$ layer-wise by fixing ${\bf\Xi}^{(r-1)}$. Then, global learning is performed for ${\bf\Xi}^{(r)}$.

{\emph{Remark 2:}} In Section \ref{sec:NetArchitecture}, we introduced the hyperparameter $\kappa$ to  control the smoothness of the $\textnormal{tanh} (\kappa{\bf x})$ function. With $\kappa\rightarrow\infty$, the function converges to the sign function. However, selecting a large $\kappa$ in training leads to \emph{gradient vanishing}, which in turn results in very slow convergence. To combat this problem, we use the continuation method\cite{allgower2012numerical,cao2017hashnet,nguyen2018learning}: we set $\kappa$ small at the beginning of the training process and increase it gradually with epoches.

\section{Application of DeepFPC in DOA Estimation}
\label{sec:DOAestimation}
We now deploy DeepFPC to address the DOA estimation problem, a typical sparse recovery problem\cite{malioutov2005sparse,xenaki2014compressive}. Applying 1-bit CS to this problem has the following advantages. Firstly, as the sensor number and array size keep increasing, 1-bit CS reduces hardware complexity and cost. Secondly, it alleviates the need for accurate calibration of all sensors, inconsistencies in which degrade the performance of alternative methods\cite{340779}.

The 1-bit measuring model for the DOA estimation problem at the $t-$th snapshot can be expressed as\cite{8555529}
\begin{equation}\label{10}
\mathbf{z}(t)={\mathcal Q}\left(\mathbf{\Lambda}\mathbf{s}(t)+\mathbf{n}(t)\right).
\end{equation}
where $\mathbf{z}(t)\in \mathbb{C}^{M}$ is measured by an $M$-element uniform linear array (ULA), $\mathbf{\Lambda}\in \mathbb{C}^{M\times N}$ is the extended steering matrix, and $\mathbf{s}(t)\in \mathbb{C}^N$ denotes the signal waveforms on the pre-defined grid of many candidate directions. Only $K\:(K\ll N)$ entries are non-zeros in $\mathbf{s}(t)$, corresponding to the $K$ narrowband far-field signals impinge onto the ULA. ${\mathcal Q}(\cdot)$ denotes the sign function operated on the complex domain, i.e., ${\mathcal Q}(\varsigma)=\operatorname{sign}(\Re\{\varsigma\})+j \cdot \operatorname{sign}(\Im\{\varsigma\})$. The complex measuring process can be converted to the real domain as
\begin{equation}\label{13}
\tilde{\mathbf{z}}(t)=\operatorname{sign}\left(\tilde{\mathbf{\Lambda}}\tilde{\mathbf{s}}(t)+\tilde{\mathbf{n}}(t)\right)
\end{equation}
where
\begin{equation*}
  \begin{split}
    \tilde{\mathbf{z}}(t)=\left[\begin{array}{c}
{\Re\{\mathbf{z}(t)\}} \\
{\Im\{\mathbf{z}(t)\}}\end{array}\right],~ & \tilde{\mathbf{\Lambda}}=\left[\begin{array}{cc}
{\Re\{\mathbf{\Lambda}\}} & {-\Im\{\mathbf{\Lambda}\}} \\ {\Im\{\mathbf{\Lambda}\}} & {\Re\{\mathbf{\Lambda}\}}
\end{array}\right], \\
     \tilde{\mathbf{s}}(t)=\left[\begin{array}{c}
{\Re\{\mathbf{s}(t)\}} \\
{\Im\{\mathbf{s}(t)\}}
\end{array}\right],~ & \tilde{\mathbf{n}}(t)=\left[\begin{array}{c}
{\Re\{\mathbf{n}(t)\}} \\
{\Im\{\mathbf{n}(t)\}}
\end{array}\right].
  \end{split}
\end{equation*}

As the one-snapshot model in \eqref{13} has become an 1-bit CS problem, we use the DeepFPC to recover the original vector $\tilde{\mathbf{s}}(t)$ from 1-bit measurements $\tilde{\mathbf{z}}(t)$ and determine the DOA.

Alternatively, in the multiple-snapshot case, we can write
\begin{equation}\label{14}
\tilde{\mathbf{Z}}=\operatorname{sign}\left(\tilde{\mathbf{\Lambda}}\tilde{\mathbf{S}}+\tilde{\mathbf{N}}\right),
\end{equation}
where $\tilde{\mathbf{Z}}\in \mathbb{R}^{2M\times L}$, $L$ is the number of snapshots. $\tilde{\mathbf{N}}\in \mathbb{R}^{2M\times L}$, and $\mathbf{S}\in \mathbb{C}^{2N\times L}$ consists of $2N$ jointly $2K$ sparse vectors, i.e., $2K$ nonzero rows. With this model, we aim to solve the following problem to estimate the DOA \cite{7227037,8555529}:

\begin{equation}\label{15}
\begin{array}{l}
{\min _{\tilde{\mathbf{S}}}\|\tilde{\mathbf{S}}\|_{1,1}+\lambda\left\|\max\left\{-\left(\tilde{\mathbf{Z}} \odot \tilde{\mathbf{\Lambda}} \tilde{\mathbf{S}}\right),\bm{0}\right\}\right\|_{1,1}} \\ {\text { s.t. }\left\|\tilde{\mathbf{s}}{(t)}\right\|_{2}=1, \quad t=1,2, \cdots, L},\end{array}
\end{equation}
where $\odot$ denotes the element-wise product and $\|\tilde{\mathbf{S}}\|_{p, q}$ is the mixed $\ell_{p,q}$-norm\cite{8555529}. The objective function can be written as $\sum_{t=1}^{L}\|\tilde{\mathbf{s}}(t)\|_{1}+\lambda\|\max \{-(\tilde{\mathbf{z}}(t) \odot \tilde{\Lambda} \tilde{\mathbf{s}}(t)), \mathbf{0}\}\|_{1}$, which means that solving the multi-snapshot model is equivalent to solving each single snapshot problem indexed by $t$ independently. Namely, DeepFPC can be trained using single measurement vectors and then applied to the multi-snapshot case.

\section{Experimental Results}
\label{sec:Experiments}
\subsection{Sparse Signal Reconstruction}
\label{sec:SignalReconstructionExperimentalResults}
We first assess the performance of DeepFPC in sparse signal recovery. The model is implemented in Python (using TensorFlow\cite{abadi2016tensorflow}) and trained with the ADAM optimizer\cite{kingma2014adam} using an exponentially decaying step-size. We consider a set of 1000 pairs $\{{\bf y}^d,{\bf x}^d\}_{d=1}^{1000}$ for training and a different set of 1000 pairs for testing. We assume that the signal dimension is $N=500$ and the sparsity level is $K=25$. The support locations are randomly chosen from the uniform distribution, and their values are drawn \textit{i.i.d.} from the standard normal. The measuring matrix $\bf \Phi$ is randomly drawn from a zero-mean Gaussian distribution with variance $1/M$. The measurement vectors have a dimension of $M=1000$ and are generated according to \eqref{1}. The normalized mean square error (NMSE), defined as $\mathrm{NMSE}\triangleq\left\|{\bf{x}}^\star-{\bf{x}}\right\|_{2}^{2} /\left\|{\bf x}\right\|_{2}^{2}$, with ${\bf{x}}^\star$ being the reconstructed vector, is used as the performance metric.

\begin{figure}
  \centering
  \includegraphics[width=2in]{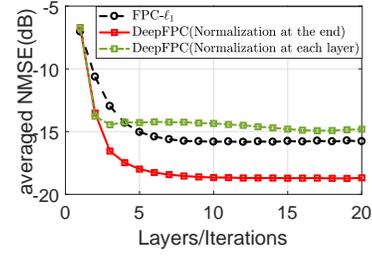}
  \caption{Recovery performance comparison between FPC-$\ell_1$ and DeepFPC in terms of averaged NMSE versus number of layers/iterations.}\label{Fig3}\vspace{-0.5em}
\end{figure}

Fig. \ref{Fig3} shows the performance comparison between DeepFPC and the FPC-$\ell_1$ algorithm in terms of the averaged NMSE for a different number of layers/iterations.
Two versions of DeepFPC are considered: one with normalization performed only at the final layer and one with normalization performed per layer. It is clear that the former model significantly improves performance; therefore, we consider this design choice in what follows. Importantly, DeepFPC (with normalization at the final layer) systematically outperforms the FPC-$\ell_1$ regardless of the number of layers/iteration, with the converged result of DeepFPC being better than the original algorithm with a 2dB margin.

We also conduct an experiment to determine whether the parameters $\bf A, B, C$ and $\bm \nu$ should be tied or not. Three 20-layer networks are trained with: (i) tied $\bf A, B, C$ and untied $\bm \nu$; (ii) untied $\bf A, B, C, \bm \nu$; and (iii) tied $\bf A, B, C, \bm \nu$. The training time and recovery error for these networks are reported in Table~\ref{tab:aStrangeTable}. We see that allowing $\nu_r$ to be separately learned per layer $r=1,\dots,R$ but maintaining $\bf A, B, C$ shared across layers strikes the best trade-off between computational complexity and recovery performance.

\begin{table}[t]
\centering
\caption{Training time and recovery error of a 20-layer network.}\label{tab:aStrangeTable}
\begin{tabular}{ccc}
\toprule  
Type& Training time [s]& Performance [dB]\\
\midrule  
tied $\bf A, B, C$, untied $\bf \nu$& 10218.15& -18.63\\
untied $\bf A, B, C$, untied $\bf \nu$& 10452.01& -18.42\\
tied $\bf A, B, C$, tied $\bf \nu$& 10194.81& -15.34\\
\bottomrule 
\end{tabular}
\end{table}

\subsection{DOA Estimation}
In this section, we evaluate the performance of DeepFPC in the task of DOA estimation in comparison to recent benchmarks, namely, the FPC-$\ell_1$\cite{8555529} and 1-bit MUSIC\cite{8700277} algorithms. 1-bit MUSIC has achieved state-of-the-art performance but assumes prior knowledge on the number of targets; this knowledge---which might not be available in a real-life setup---is not required by FPC-$\ell_1$ and the proposed DeepFPC.

We consider an 8-layer DeepFPC, since the performance improves slowly over this number of layers as shown in Fig.~\ref{Fig3}. DeepFPC is trained using a training set of 1000 signal-measurement pairs, generated according to \eqref{10}. The DOAs in the training set are drawn from the uniform distribution and their number varies from 2 to 10. In \cite{8555529} and \cite{7227037} the target signals are assumed to be quadrature phase shift keying (QPSK) modulated, and the sampling grid is carefully designed across equally spaced spatial frequencies so as to reduce coherence between the steering matrix columns. In this experiment, we consider a more challenging scenario: the target signals are drawn from the standard normal and the grid is formed by uniformly spaced sampling with $1^{\circ}$ (i.e., $N=180$), where the number of sensors is set to 40.

We employ a testing set of 500 pairs, considering $K=6$ (i.e., the number of narrowband far-field signals impinge onto the ULA) with DOAs set to $[-40^{\circ}$,$-16.7^{\circ}$,$-4.2^{\circ}$, $1.6^{\circ}$, $15.7^{\circ}$, $60^{\circ}]$. The test signals are drawn from the standard normal distribution. Unlike the training set, the measurements in the testing set are contaminated by \textit{i.i.d.} Gaussian noise. The signal-to-noise ratio (SNR) in dB is defined as $\mathrm{S N R}=10 \log _{10}\left(\sigma_{s}^{2} / \sigma_{n}^{2}\right)$, where $\sigma_{s}^{2}$ and $\sigma_{n}^{2}$ are respectively the variances for the signal and the noise. The results for different methods are averaged over $R=500$ Monte Carlo runs. The mean absolute error (MAE) is used as metric to evaluate the DOA estimation performance:
$\mathrm{MAE}=\frac{1}{J K} \sum_{j=1}^{J} \sum_{k=1}^{K}\left|\hat{\alpha}_{k, j}-\alpha_{k}\right|$,
where $\alpha_{k, j}$ denotes the $k$-th DOA estimate in the $j$-th run, and $\alpha_{k}$ is the true DOA of the $k$-th signal. The FPC-$\ell_1$ algorithm\cite{8555529} operates in an outer-inner iteration loop. In the inner loop, it recovers the sparse vector with a fixed regularization parameter $\lambda_i$, and in the outer loop, $\lambda_i$ is updated as $\lambda_i=c\lambda_{i-1}$. Here, we set $c=\lambda_0=1.1$, the step size $\tau=0.01$, and the numbers of inner and outer iterations are 200 and 20, respectively.

\begin{figure}
  \includegraphics[width=3.5in]{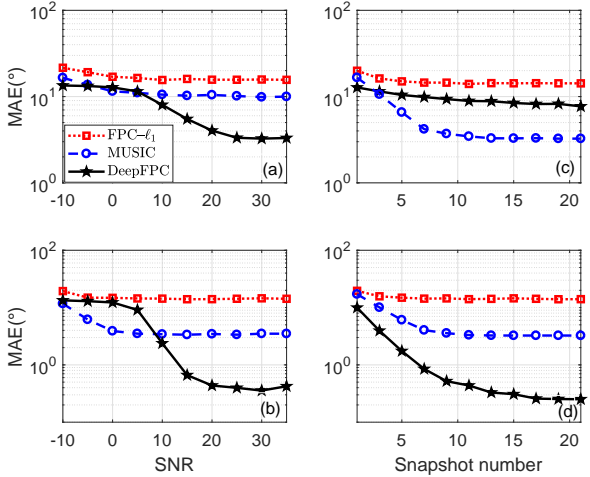}\vspace{-0.5em}
  \caption{DOA estimation error for three methods ($M=40$): (a) MAE versus SNR with snapshot number $L=3$; (b) MAE versus SNR with snapshot number $L=10$; (c) MAE versus snapshot number $L$ with SNR=$5$dB and (d) MAE versus snapshot number $L$ with SNR=$20$dB.}\label{Fig4}\vspace{-0.5em}
\end{figure}

Fig. \ref{Fig4}(a),(b) plot the MAE versus the SNR for 3 and 10 snapshots, respectively, whereas Fig. \ref{Fig4}(c),(d) depict the MAE versus the number of snapshots for SNR values of 5dB and 20dB, respectively.
The results show that, even though DeepFPC stems from FPC-$\ell_1$, it systematically outperforms the latter independent of the SNR level or the number of snapshots. Furthermore, DeepFPC performs significantly better than MUSIC at medium to high SNRs without requiring prior knowledge on the number of targets. It is worth observing that, in a high SNR environment (20dB), the MAE achieved by DeepFPC is an order of magnitude lower than that of MUSIC [see Fig. \ref{Fig4}(d)]. DeepFPC is trained using noise-free signals; hence, in low SNRs, it is outperformed by MUSIC. We experimentally found that the robustness of DeepFPC against high-level noise is not effectively improved by using noisy signals for training. Hence, improving the DOA estimation accuracy of DeepFPC in the low SNR regime is an open problem, which we plan to address in subsequent work.

\begin{figure}
  \centering
  \includegraphics[width=2.2in]{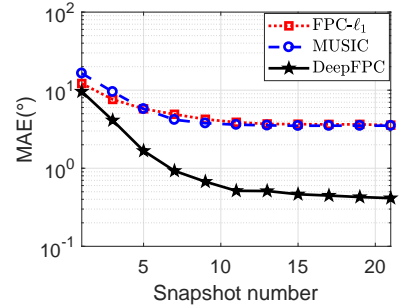}\vspace{-0.5em}
  \caption{Comparisons of MAE versus snapshot numbers between FPC-$\ell_1$ and DeepFPC for the optimized and uniform grid (SNR = 20dB).}\label{Fig6}\vspace{-0.5em}
\end{figure}

In the previous simulation, the grid was uniformly spaced, without considering the coherence between the columns of the steering matrix $\bf\Lambda$. When the grid is designed such that $\theta_{i}=\sin ^{-1} [2i /M]$, $i=0,1,\dots,M-1$, the columns of~$\bf\Lambda$ form an orthogonal system, which helps improving the recovery performance of conventional algorithms like FPC-$\ell_1$ \cite{doi:10.1121/1.4883360}. Fig.~\ref{Fig6} depicts the MAE performance of the different approaches versus the snapshot number when the grid is optimized and the SNR is set to 20dB. Contrasting these results with those in Fig.~\ref{Fig4}(d)---which depicts results with a uniform grid---, we corroborate that the FPC-$\ell_1$ algorithm achieves better performance, which is equivalent to that of MUSIC. We also observe that the performance of DeepFPC is not notably influenced by the grid structure and remains significantly above the performance of FPC-$\ell_1$ and MUSIC.

To sum up, the proposed DeepFPC outperforms the conventional FPC-$\ell_1$ algorithm for sparse signal recovery. In the task of DOA estimation, although its performance is relatively limited at low SNRs, DeepFPC performs significantly better than MUSIC and FPC-$\ell_1$ at medium and high SNRs. Moreover, unlike MUSIC, DeepFPC does not require information about the target number, and it is less sensitive than FPC-$\ell_1$ to the grid setting.

\section{Conclusion}
\label{sec:conclusion}
We proposed a novel deep neural network called DeepFPC, designed by unfolding the FPC-$\ell_1$ recovery algorithm for 1-bit CS. We showed that, once properly trained, DeepFPC delivers much higher sparse signal recovery performance than the basic algorithm. We also demonstrated the capacity of DeepFPC in the task of DOA estimation. Experimentation has shown that DeepFPC systematically outperforms FPC-$\ell_1$ and performs better than the one-bit MUSIC algorithm for medium and high SNRs. DeepFPC can be applied to other 1-bit CS problems in the domains of image processing, channel estimation and localization.\vspace{-0.5em}

%
%

\ifCLASSOPTIONcaptionsoff
  \newpage
\fi



%

\bibliographystyle{IEEEtran}


%








\end{document}